%% file: conference_101719.tex
\documentclass[conference]{IEEEconf}
\IEEEoverridecommandlockouts
\usepackage{float}
\usepackage{cite}
\usepackage{booktabs}
\usepackage{amsmath,amssymb,amsfonts}
\usepackage{algorithmic}
\usepackage{graphicx}
\usepackage{textcomp}
\usepackage{comment}
\usepackage{fixltx2e}
\usepackage{ragged2e}
\usepackage{subcaption}
\usepackage{gensymb}
\usepackage{hyperref}
\usepackage{soul}
\usepackage{cellspace} 
\usepackage{multirow}
\usepackage[table]{xcolor}
\usepackage[acronym]{glossaries}
\loadglsentries{glossaries}

\def\BibTeX{{\rm B\kern-.05em{\sc i\kern-.025em b}\kern-.08em
    T\kern-.1667em\lower.7ex\hbox{E}\kern-.125emX}}

\begin{document}

\title{Self-Supervised Multimodal NeRF for Autonomous Driving

\thanks{The author is with the AVL Software and Functions GmbH, Germany and Technische Hochschule Deggendorf, Germany 
        {\tt\small\ gaurav.sharma@avl.com$^{1}$, ravi.kothari@avl.com$^{2}$, josef.schmid@th-deg.de$^{3}$}}
}
%1\textsuperscript{st}
\author{
Gaurav Sharma$^{1}$, Ravi Kothari$^{2}$, Josef Schmid$^{3}$
}

\maketitle

\begin{abstract}
In this paper, we propose a \gls{nerf} based framework, referred to as \gls{nvsf}. It jointly learns the implicit neural representation of space and time-varying scene for both LiDAR and Camera. We test this on a real-world  autonomous driving scenario containing both static and dynamic scenes. Compared to existing multimodal dynamic \gls{nerf}s, our framework is self-supervised, thus eliminating the need for 3D labels. For efficient training and faster convergence, we introduce heuristic-based image pixel sampling to focus on pixels with rich information. To preserve the local features of LiDAR points, a Double Gradient based mask is employed. Extensive experiments on the KITTI-360 dataset show that, compared to the baseline models, our framework has reported best performance on both LiDAR and Camera domain. Code of the model is available at https://github.com/gaurav00700/Selfsupervised-NVSF
% \href{https://github.com/xxx}{Github}. 

\end{abstract}

\section{INTRODUCTION}
\gls{lidar} and Camera represent two of the most critical sensors in the perception systems of \gls{av}. However, the collection and annotation of such data is an inherently labor-intensive and costly endeavor, necessitating substantial manual effort and resources. The generation of highly accurate synthetic data offers a promising alternative, potentially circumventing these challenges. There are many model-driven approaches available for generating novel view synthetic \gls{lidar} and Camera data based on scene formulation, such as Game-engine-based \cite{dosovitskiy2017carla}, Explicit-model-based \cite{manivasagam2020lidarsim} and Implicit model based \cite{mildenhall2021nerf}. All existing methods come with their own set of limitations. For example, CARLA\cite{dosovitskiy2017carla} operates within a hand-made environment, while some approaches fail to construct view-dependent attributes such as raydrop and point cloud intensity or rely on explicit scene models. As a result, these methods struggle to generalise effectively, particularly in the \gls{lidar} domain, due to the challenges in accurately mapping the physics of active sensors. Achieving accurate and realistic \gls{nvs} is also crucial for minimising the sim-to-real domain gap.

\begin{figure} [h!]
  \centering
  \includegraphics[width=0.48\textwidth]{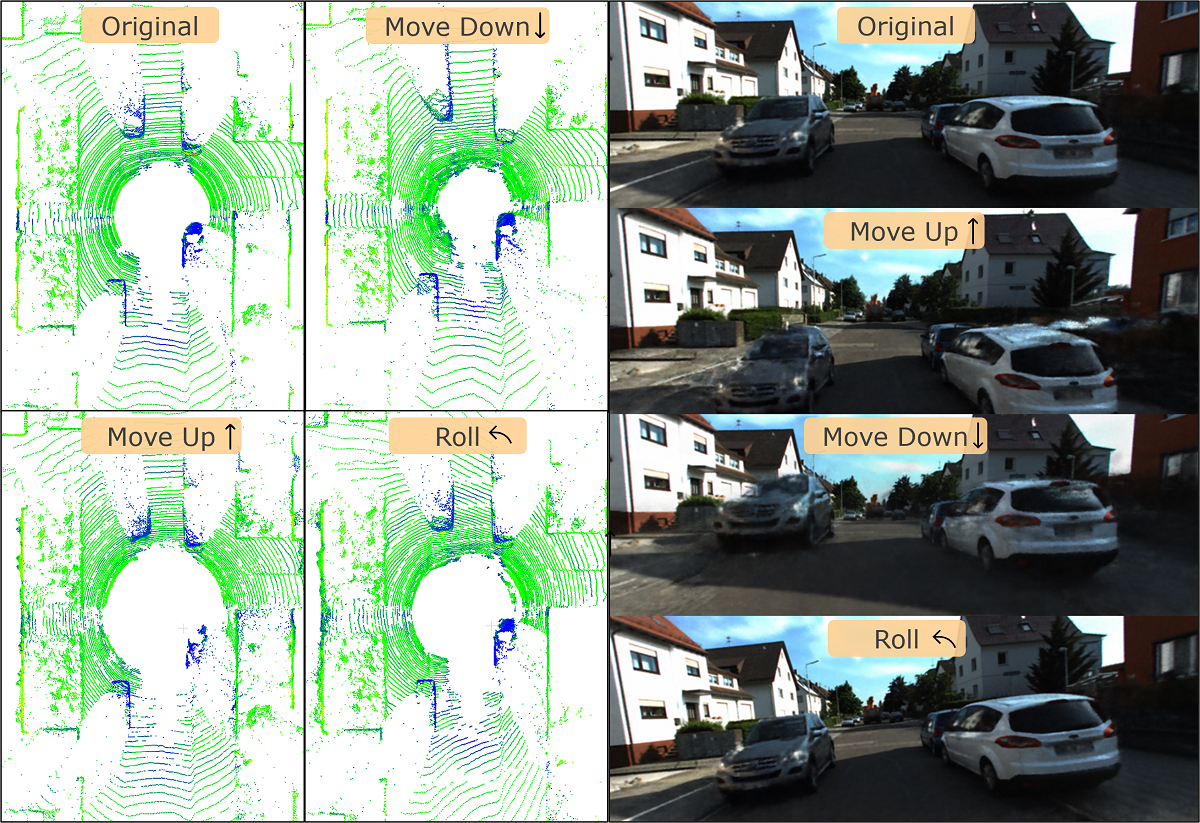}
  \caption{Joint Novel View Synthesis for Camera and LiDAR sensor by the proposed method. Along with extrinsic attributes, intrinsic sensor parameters can be also varied.}
  \label{fig: NVS Lidar and Camera}
\end{figure}

Advancements in differential volume rendering and \gls{nerf} based methods have ushered in new possibilities in \gls{nvs}. Thus, we propose the \gls{nvsf} framework, leveraging \gls{nerf} to synthesise novel views for both \gls{lidar} and Camera. Central to the proposed framework is joint learning of implicit neural representation of scene geometry and view-dependent attributes for both \gls{lidar} and Camera sensors. Exploration of \gls{lidar} domain adaptation through the synthesis of \gls{pcd} using \gls{nerf} spans both static and dynamic scenes, with the latter posing unique challenges due to \gls{nerf}'s inherent operating principles. To address the dynamic scene problem, the proposed framework draws inspiration from LiDAR4D \cite{zheng2024lidar4d} and K-Planes \cite{fridovich2023k}, which factorise a canonical scene, 4D volume of space and time through multiple 2D planes. It decomposes a canonical Scene in space and time, thereby making the model interpretable between static and dynamic objects. Unlike \gls{nsg} \cite{ost2021neural} based methods, the proposed framework does not require any external supervision from 3D annotation to decompose a canonical scene into corresponding foreground and background scenes. To resolve this K-Planes based network handles dynamic scene of both non-rigid and rigid objects such as pedestrians, cars, trucks, etc. The proposed framework empowers the novel view synthesis of \gls{lidar} and Camera in real-world driving scenarios of \gls{av}s, supporting diverse applications such as minimising domain gaps, domain invariant training, experimenting with different sensor specifications, sensor configurations on \gls{av} and revival of corrupted data.

Our contributions are threefold:
\begin{enumerate} 
    \item  We propose a multimodal self-supervised \gls{nvs} framework which jointly learns the implicit Neural representation of a 4D spatio-temporal scene for \gls{lidar} and Camera 
    \item We introduce a Heuristic pixel sampling method using Multinomial Probability Distribution based on reconstruction error of image pixels during training
    \item To improve feature alignment in synthesised \gls{lidar} point clouds, we propose a Double Gradient-based clipping mask for capturing sparse features
\end{enumerate}

\section{RELATED WORK}
\label{sec2}

\textit{(a) NeRF:} NeRF \cite{mildenhall2021nerf} and its derivatives \cite{ barron2022mip, barron2023zip} have demonstrated remarkable performance in 3D view reconstruction using multi-view 2D images. Conventional \gls{nerf} relies on \gls{mlp} to learn the scene content at each queried point, but the need to compute thousands of points along a ray significantly slows down the learning process. Instant-NGP (iNGP) \cite{muller2022instant}, TensoRF \cite{chen2022tensorf} and K-Planes \cite{fridovich2023k} employ learnable embeddings or voxels to represent 3D scenes, which significantly reduce the computational burden of the rendering \gls{mlp}. Recent research efforts centred on \gls{nerf}s have driven significant advancements and notable successes in \gls{nvs} tasks. A diverse range of neural representations, including \gls{mlp}s \cite{mildenhall2021nerf, barron2022mip, barron2023zip}, voxel grids \cite{fridovich2022plenoxels, sun2022direct}, tri-planes \cite{chan2022efficient, hu2023tri}, vector decomposition \cite{chen2022tensorf}, and multi-level hash grids \cite{muller2022instant}, have been extensively explored for both reconstruction and synthesis. However, the majority of these studies are concentrated on object-centric reconstructions of small indoor scenes. More recently, several works \cite{sun2024dil, tancik2022block, tao2023lidar, zhang2023nerf, zheng2024lidar4d, rematas2022urban, wang2023neural} have gradually expanded these approaches to large-scale outdoor environments. However, most existing approaches focus on either the Camera or \gls{lidar} domain, with only a few \cite{tonderski2023neurad, yang2023unisim} addressing both modalities concurrently. A significant challenge remains in handling both static and dynamic scenes, a crucial requirement for real-world driving scenarios. This limitation impacts the scalability of such networks. In this paper, we propose a unified framework that offers \gls{nvs} for both \gls{lidar} and Camera. 

\hspace{0.4cm}

\textit{(b) \gls{nerf} for LiDAR:} Traditional \gls{nerf} models were initially designed for Camera images and cannot be directly applied to \gls{lidar} point clouds due to the fundamental structural differences between these two data types. Unlike images, which capture dense 2D pixel grids, \gls{lidar} provides sparse, 3D point cloud data, making it challenging to adapt \gls{nerf} for \gls{lidar}'s spatial properties. Numerous approaches have been developed to adapt \gls{nerf} to the \gls{lidar} domain. Early notable works include Neural \gls{lidar} Fields for Novel view synthesis \cite{huang2023neural} and LiDAR-NeRF \cite{tao2023lidar}, which can synthesise novel synthesis of \gls{lidar} point cloud. These were followed by additional research, such as NeRF-LiDAR \cite{zhang2024nerf}, NeuRAD \cite{tonderski2023neurad} and LiDAR4D \cite{zheng2024lidar4d}. Among these, NeRF-LiDAR \cite{zhang2024nerf} and NeuRAD \cite{tonderski2023neurad} stand out for utilising both \gls{lidar} and Camera data jointly to learn \gls{nerf} fields, enhancing their ability to generate more accurate and detailed reconstructions by leveraging the complementary nature of both sensor modalities. The integration of \gls{lidar} and Camera data mitigates the inherent limitations associated with single-modal approaches, thereby enabling more robust and effective neural radiance field implementations in complex 3D environments. However, both of these models utilise 3D annotation of \gls{lidar} point cloud for tracking moving objects. As data annotation is an expensive process therefore, it is not always possible to invest time and resources in it. In our work, we are learning dynamic scene content in a self-supervised manner by employing the scene flow module. 

\section{METHOD}
\label{sec4}

\begin{figure*}[h!]
  \centering
  \includegraphics[width=1\textwidth]{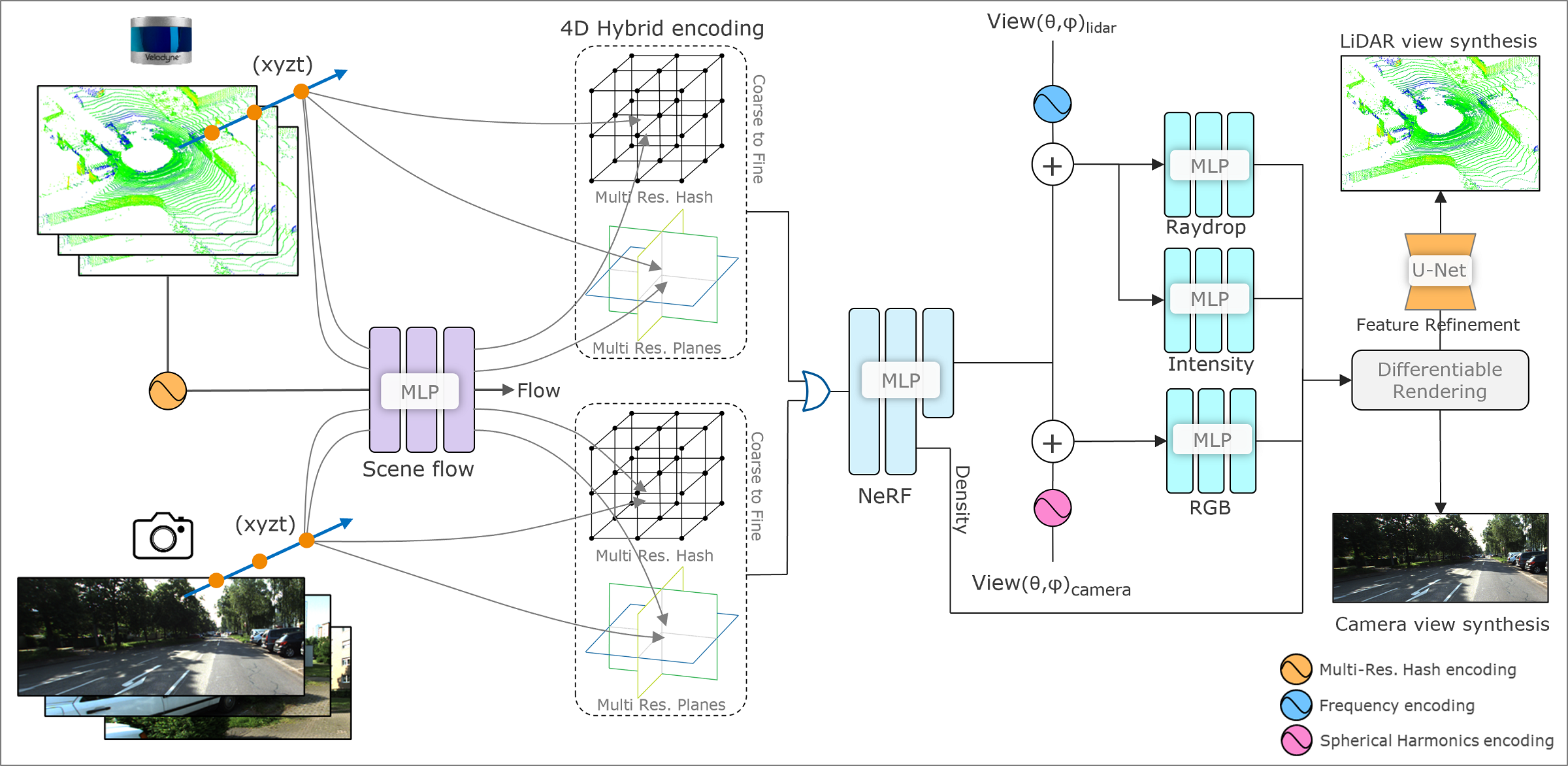}
  \caption{The proposed network consists of Scene flow \gls{mlp} for learning the temporal features, 4D hybrid hash-planes encoding for the encoding of space and time coordinates, and a shared \gls{mlp} for learning the scene density and global features. A differentiable rendering module for calculating depth, intensity, raydrop of \gls{lidar} point cloud and pixel colour. U-Net network is used for global refinement of \gls{lidar} renderings, as mentioned in LiDAR4D\cite{zheng2024lidar4d}.}
  \label{fig: proposed nvsf network}
\end{figure*}

The network of the proposed framework is inspired by existing works such as NeRF\cite{mildenhall2021nerf}, LiDAR-NeRF\cite{tao2023lidar}, K-Planes\cite{fridovich2023k}, LiDAR4D\cite{zheng2024lidar4d} and Neural Scene Flow Prior\cite{NEURIPS2021_41263b9a}. A novel contribution of the proposed network includes multimodal training by using both \gls{lidar} and Camera data for jointly learning \gls{nerf} models of \gls{lidar} and Camera, heuristic image pixel sampling, and improved gradient loss for better feature alignment. Figure \ref{fig: proposed nvsf network} illustrates the proposed network; it comprises five individual \gls{mlp}s and two parameterised 4D hybrid encodings using Multi-resolution Hash encoding \cite{muller2022instant} and K-Planes encoding\cite{fridovich2023k}. Each module of the proposed network serves a different learning objective such as: \textbf{(1)} Scene Flow \gls{mlp} learns the forward and backward scene flow; \textbf{(2)} Positional, Spherical Harmonics and Multi-Resolution Hash encoding encodes positional and temporal information into high-dimensional feature vectors; \textbf{(3)} 4D Hybrid encoding encodes the space and time information using multi-resolution hash and 2D-planes for the decomposition of static and dynamic contents of a scene; \textbf{(4)} NeRF \gls{mlp} learns the implicit neural representation of 4D scene in spatio-temporal coordinate frame; \textbf{(5)} Three radiance \gls{mlp} handles the view-dependent attributes such as raydrop, intensity of \gls{lidar} point cloud and RGB pixel value of Camera image; \textbf{(6)} U-Net is used for global refinement of raydrop for better consistency of synthesised point cloud throughout the scene.

By combining all modules, there are around 132 million trainable parameters to be optimised during network training. Unlike recent research on NeRF-based multimodal frameworks for \gls{nvs} such as MARS\cite{wu2023mars}, NeRF-LiDAR\cite{zhang2023nerf} and NeuRAD\cite{tonderski2023neurad}, the proposed network decouples static (background) and dynamic (foreground) scene components without any external supervision. All modules work coherently to synthesise views of \gls{lidar} and Camera with high fidelity for large-scale static and dynamic scenarios of autonomous vehicle.  Therefore, the proposed \gls{nvsf} of \gls{lidar} and Camera can be formalised as a function of \gls{lidar} $(\sigma,\mathbf{i},\mathbf{p})=f_{lidar}(\mathbf{x},\theta)$ and Camera $(\sigma,\mathbf{c})=f_{Camera}(\mathbf{x},\theta)$. Where $\sigma$, $\mathbf{i}$, $\mathbf{p}$, $\mathbf{c}$, $\mathbf{x}$ and $\theta$ denotes point  density, LiDARintensity, ray-drop probabilities, Camera RGB, spatial coordinates and view angles values respectively. 

\subsection{Geometric Feature Alignment}
\gls{lidar}'s uniform laser projection creates characteristic patterns in point clouds, particularly in flat regions, such as road surface, building walls and car bonnet. 
The vanilla \gls{nerf} struggles with capturing these local features as it is optimised with a global pano image objective. As a result, despite performing well on PCD metrics such as \gls{cd}, it struggles with \gls{psnr} score. LiDAR-NeRF \cite{tao2023lidar} proposed structural regularization loss, to improve local geometric consistency. 

\begin{figure}[h]
    \subfloat[Single gradient mask]{%
        \includegraphics[width=.49\linewidth]{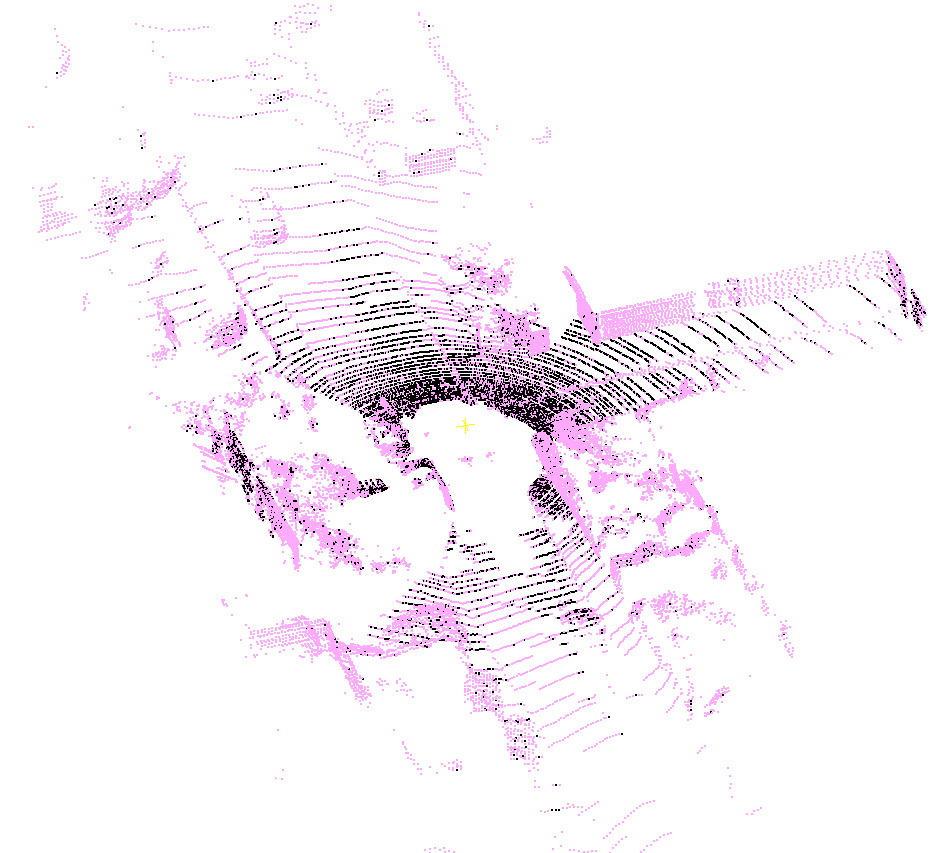}
        \label{fig: single gradient loss}%
    }\hfill
    \subfloat[Double gradient mask]{%
        \includegraphics[width=.49\linewidth]{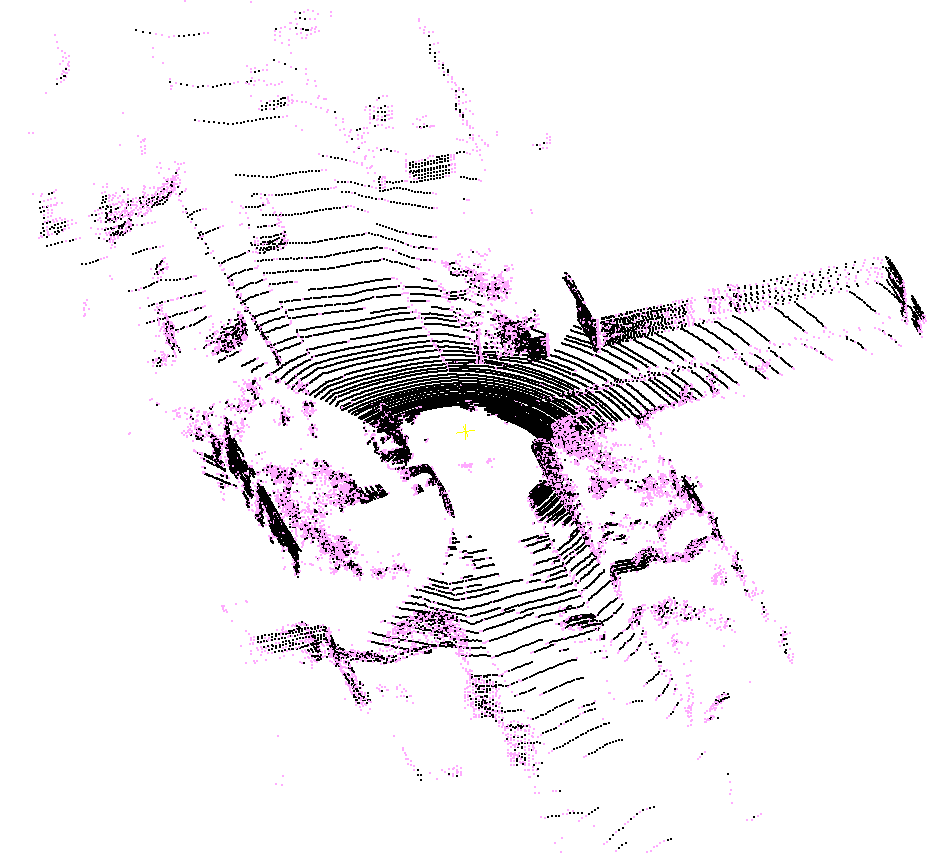}%
        \label{fig: double gradient loss}%
    }
    \caption{Compared to a single-gradient approach (a), our method (b) captures sparse features. The black highlighted regions are chosen for structural regularization, while the pink-highlighted areas are skipped due to their lack of geometric features.}
    \label{fig: single vs double gradient loss}
\end{figure}

To compute targeted areas for applying gradient-based loss, single gradients of the ground truth panoramic depth image were used to create the clipping mask. A threshold was applied to these gradients to control the coverage of the gradient loss through the mask. However, analysis revealed that the single gradient-based clipping mask failed to capture distant regions of the point cloud, as illustrated in Figure \ref{fig: single gradient loss}. Increasing the threshold for the clipping mask led to over-smoothing of the panoramic depth image. As gradient loss was applied to non-flat areas, reducing the accuracy of the synthesised point cloud. This issue also affects LiDAR4D \cite{zheng2024lidar4d}, which uses the same gradient loss function as evident in their codebase.

Unlike single gradients, double gradients of pano depth image of far away regions such as ground and walls of the scene are zero. Gradient loss can be computed on a large relevant area of point cloud by using a small threshold on pano image, as shown in Figure \ref{fig: double gradient loss}. This method covers both sparse-texture (far distance) and dense-texture (near distance) regions of the point cloud. Figure \ref{fig: double gradient loss} shows the points where Gradient loss is computed. The double gradient mask is better as it can cover more (black region) geometrically structured regions for feature alignment. 

% \begin{figure} [h!]
%   \centering
%   \includegraphics[width=0.3\textwidth]{Bilder/pcd_grad_x.png}
%   \caption{Point cloud with Gradient loss through single gradient mask. Black and Pink colour points are points with and without gradient loss respectively.}
%   \label{fig: single gradient loss}
% \end{figure}
% \begin{figure} [h!]
%   \centering
%   \includegraphics[width=0.3\textwidth]{Bilder/pcd_grad_xx.png}
%   \caption{Point cloud with Gradient loss through double gradient mask. Black and Pink colour points are points with and without gradient loss respectively.}
%   \label{fig: double gradient loss}
% \end{figure}

%showing figure in 2x2 grid

Gradient loss through a Double gradient-based clipping mask is calculated at small cropped regions (patches) of the pano depth image. Combined gradient loss for both $x$ and $y$ directions are optimised in local patches. Gradient Loss is given by the Equation \ref{eq: gradient loss}, where $D$ and $\hat{D}$ is ground truth and predicted depth of training rays in local patches,  $\nabla = D_{i,j} - D_{i,j+1}$ is single gradient (first-order),  $ M^{(2)}$ is double gradient (second-order) mask local path in pano depth image , as given by the Equation \ref{eq: double gradient mask} and $\tau $ is double gradient threshold.

% where $R$ is the set of training rays of local patches, $G_M(\cdot)$ and $\hat{G}_{M}(\cdot)$ denotes the gradient operation of ground truth and prediction in $x$ and $y$ directions respectively. The loss equation is derived from  LiDAR-NeRF \cite{tao2023lidar} structural regularization.

\begin{equation} \label{eq: gradient loss}
\mathcal{L}_{\text{grad}} = 
\left\| M^{(2)} \cdot \left( \nabla \hat{D} - \nabla D \right) \right\|_1 
\end{equation}
\vspace{-0.5em}
\begin{align} \label{eq: double gradient mask}
    M^{(2)} &= \begin{cases}
    1 & \text{if } \left|| \nabla D(i,j) \right| - \left| \nabla D(i,j+1) \right|| < \tau \\
    0 & \text{otherwise}
    \end{cases}
\end{align}

% \begin{align}
% G_x(D) &= D(i, j) - D(i, j+1) \\
% G_y(D) &= D(i, j) - D(i+1, j) \\
% \end{align}

% \begin{align}
% G_{xx}(D) &= |G_x(D)(i, j)| - |G_x(D)(i, j+1)| \\
% G_{yy}(D) &= |G_y(D)(i, j)| - |G_y(D)(i+1, j)|
% \end{align}

% \begin{align}
% \mathcal{L}_{\text{grad}} &= \mathcal{L}_{\text{grad}}^x + \mathcal{L}_{\text{grad}}^y \\
% \mathcal{L}_{\text{grad}}^x &= \left\| M_x \cdot \left( G_x(\hat{D}) - G_x(D) \right) \right\|_1 \\
% \mathcal{L}_{\text{grad}}^y &= \left\| M_y \cdot \left( G_y(\hat{D}) - G_y(D) \right) \right\|_1
% \end{align}

% \begin{equation} \label{eq: gradient loss}
%     \mathcal{L}_{\mathrm{grad}}\;=\;\left|\left|\hat{G}_{M}(\mathbf{R})-G_{M}(\mathbf{R})\right|\right|_{1}
% \end{equation}

\subsection{Heuristic Ray Sampling}
\begin{figure} [h!]
  \centering
  \includegraphics[width=0.49\textwidth]{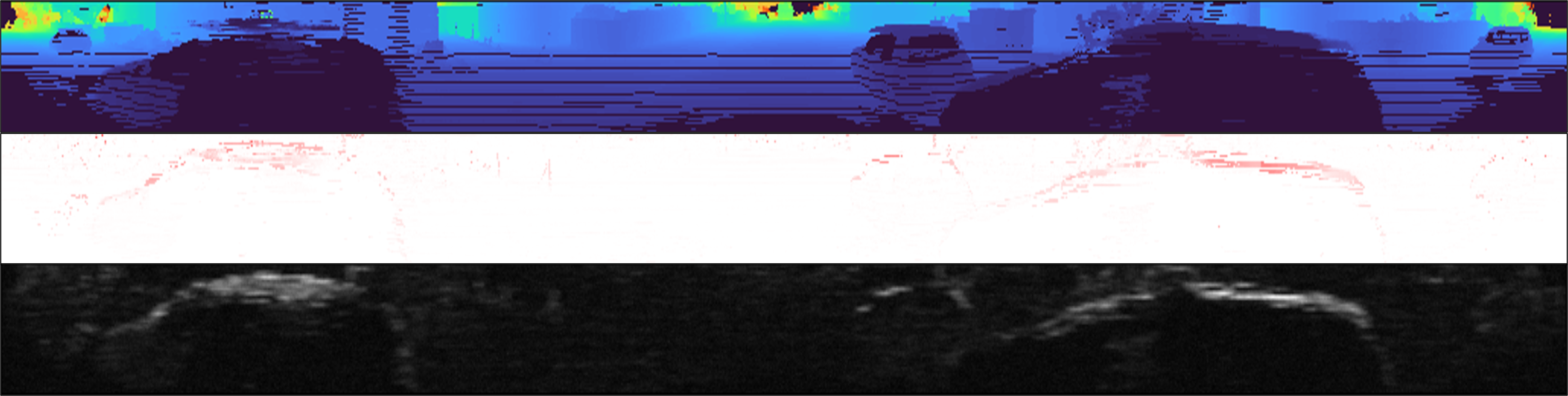}
  \caption{Figure shows the Heuristic based pixel sampling of \gls{lidar}'s Pano image. From top to bottom: Pano depth, Reconstruction loss and Heuristic pixel sampling. Red pixels in the middle image indicate high reconstruction loss, thus increasing their sampling probability, shown as white pixels in the bottom image.}
  \label{fig: comparison between Random and Hueristic based pixel sampling}
\end{figure}

Naive NeRF-based models originally used a uniform ray sampling approach, this method assumed that each pixel contained the same amount of information. However, this assumption fails in large, unbounded scenes, such as those in autonomous driving, where information entropy is non uniform. Failure to consider this leads to \gls{nerf} rendering ghost artifacts, especially around complex objects such as, pedestrians. To address this, strategy of Non-uniform sampling was introduced in \cite{otonari2022non}, but it was targeted for the small indoor 360$^{\circ}$ scenes. The \gls{nvsf} employs heuristic-based pixel sampling during training, guided by a multinomial probability distribution, as given the Equation \ref{eq: cd scene flow}. This content-aware ray sampling adapts to pixel-wise reconstruction loss, giving higher sampling probabilities to pixels with higher loss. As illustrated in Figure \ref{fig: comparison between Random and Hueristic based pixel sampling}, this approach prioritizes pixels with higher reconstruction loss compared to random sampling. Since all pixels have a positive likelihood of being sampled, this approach balances exploration and exploitation by focusing on both high-loss and unexplored regions. To introduce variability, sampled pixel indices are perturbed through jittering ($\epsilon \sim [0, 1]$). The framework alternates between heuristic and random sampling across training epochs, allowing the network to explore new pixels in one epoch and minimise reconstruction loss in the next, thereby improving focus on challenging pixels.

\subsection{Multimodal NeRF}
Depth supervision from \gls{lidar} has already shown to improve image based NeRF\cite{depthnerf}, similarly images complement the sparse \gls{lidar} data. Building on this NeRF-LiDAR \cite{zhang2024nerf} and NeuRAD \cite{tonderski2023neurad} also use mutlimodal data for \gls{nvs} of both sensors. But these networks require 3D annotations for tackling dynamic problem, which involves additional labelling effort. The proposed framework can do \gls{nvs} for both sensors in a self-supervised manner by employing a 3D flow network. A strategy is required to fuse encoded features effectively for improved convergence of the proposed network. Recent research \cite{tang2024alignmif} indicates that different modalities often contradict each other due to spatial and temporal misalignment. Moreover, \gls{lidar} and Camera modalities exhibit inconsistent representations and converge at different rates, leading to uncoordinated convergence issues \cite{peng2022balanced}, \cite{sun2021learning}, \cite{ wang2020makes}. We conduct extensive experiments as shown in Table \ref{tab: encoding combinations} with various combinations of encoded features from both modalities: 
\begin{enumerate}
    \item Common encoding for Camera and LiDAR
    \item Separate encoding for Camera and LiDAR
    \item Separate encoding for Camera and LiDAR with implicit fusion before passing to NeRF \gls{mlp}
\end{enumerate}

Separate 4D Hybrid Hash-Planar encoding for \gls{lidar} and Camera data performed best among all the three options

\begin{comment}
\begin{table} [h!]
  \centering
  \includegraphics[width=0.49\textwidth]{Bilder/encoding_combinations.png}
  \caption{Comparison for different combination of 4D Hybrid Hash-Planar encoding for LiDAR and Camera. \textbf{Bold} cell values are showing best results in respective metrics}
  \label{tab: encoding combinations}
\end{table}
\end{comment}

\begin{table}[h]
\setlength{\tabcolsep}{5pt}
    \begin{tabular}{|p{2cm}|c|c|c|c|c|}
    \hline
    \arrayrulecolor{black}
         \rule{0pt}{10pt} \multirow{3}{*}{\textbf{Strategy}} & \multicolumn{3}{c|}{\textbf{LiDAR}} & \multicolumn{2}{c|}{\multirow{2}{*}{\textbf{Camera}}} \\ \cline{2-4} 
         \rule{0pt}{10pt} & \multicolumn{1}{c|}{\textbf{PCD}}& \multicolumn{2}{c|}{\textbf{Range}} & \multicolumn{2}{c|}{\textbf{}}  \\ \cline{2-6} 
         \rule{0pt}{10pt} & CD $\downarrow$ & SSIM $\uparrow$ & PSNR $\uparrow$ & SSIM $\uparrow$ & PSNR $\uparrow$ \\ \hline
         Same Enc. & 0.175 & 0.76  & 27.12   & 0.712 & 24.97  \\ \hline
         Separate Enc. & \textbf{0.110}  & \textbf{0.792}  & \textbf{27.85}  
         &  \textbf{0.811}  & \textbf{25.21}  \\ \hline
         Separate Enc.+F. & 0.161 & 0.757  & 27.52   & 0.809  & 25.14  \\ \hline
    \end{tabular}
    \caption{Comparison of different encoding strategies.}
    \label{tab: encoding combinations}
\end{table}

\subsection{Implementation}
We follow the classical differential rendering approach used in \gls{nerf} models \cite{mildenhall2021nerf, tao2023lidar, zheng2024lidar4d}. The view direction of rays is similar to the pattern of a spinning mechanical \gls{lidar}, whose beams are cast in $360^\circ{}$. The expected depth $\hat{D}({\bf r})$ of a beam can be obtained by integrating over samples given by equation \ref{eq: compositing lidar depth}, which is derived from LiDAR4D \cite{zheng2024lidar4d}.

\begin{equation} \label{eq: compositing lidar depth}
    \hat{D}({\bf r})=\sum_{i=1}^{N}T_{i}\left(1-e^{-\sigma_{i}\delta_{i}}\right)t_{i},\quad\alpha_{i}=1-e^{-\sigma_{i}\delta_{i}}
\end{equation}

Where $T_{i}=e^{\left(-\sum_{j=1}^{i-1}\sigma_{j}\delta_{j}\right)}$ indicates the accumulated transmittance along ray $\bf r$ to the sampled point $t_i$, and $\delta_{i}=\,t_{i+1}-t_{i}$ is the distance between adjacent samples.  $t_{i}$ is the depth value of queried points on the ray $r$, and $\alpha_{i}$ is the opacity value. The expected depth represents the distance from the \gls{lidar} sensor. Moreover, both the origin and viewing direction of the ray are transformed to the global world coordinate system, allowing the sampled points  $t_{i}$  to be actual points in the real world and consistent across multiple frames (pano image). The view dependent attributes for \gls{lidar} are same as in LiDAR4D \cite{zheng2024lidar4d} and for Camera attributes are calculated as per Image NeRF\cite{mildenhall2021nerf}.

%View-dependent attributes intensity and raydrop probability of LiDAR were predicted by $\text{MLP}_{intensity}$ and $\text{MLP}_{raydrop}$ at a sampled point along a ray. Alpha composition was applied to calculate final values of intensity ${\hat{\cal I}}$ and raydrop ${\hat{\cal R}}$, it is given by equations \ref{eq: final intensity} and  \ref{eq: final raydrop} respectively. Similarly, the Camera's image pixel RGB values ${\hat{\cal C}}$ were calculated using alpha compositing of point-wise predicted colour values of a ray from the Camera pixel using $\text{MLP}_{rgb}$, it is given by the equation \ref{eq: final rgb pixel}.

%\begin{equation} \label{eq: final intensity}
    %{\hat{\cal I}}({\bf r})=\sum_{i=1}^{N}{T}_{i}\alpha_{i}i_{i}
%\end{equation}
%\begin{equation} \label{eq: final raydrop}
    %{\hat{\cal P}}({\bf r})=\sum_{i=1}^{N}{T}_{i}\alpha_{i}p_{i}
%\end{equation}
%\begin{equation} \label{eq: final rgb pixel}
    %{\hat{\cal C}}({\bf r})=\sum_{i=1}^{N}{T}_{i}\alpha_{i}c_{i}
%\end{equation}

% where $i_i$, $p_i$ and $c_i$ are the point-wise intensity, ray-drop probability and RGB pixel colour outputs of $\text{MLP}_{intensity}$, $\text{MLP}_{raydrop}$ and $\text{MLP}_{rgb}$ respectively.

As vehicle motion range may span a long distance in autonomous driving scenarios, it is extremely difficult  to establish long-term correspondences in the canonical space without a scene flow network. A few models for the prediction of points translation in a space-time volume are Neural scene flow prior \cite{li2021neural},  Fast Neural Scene Flow (FastNSF) \cite{Li_2023_ICCV},  NeuralPCI  \cite{zheng2023neuralpci}, Dynibar\cite{li2023dynibar} and LiDAR4D \cite{zheng2024lidar4d}. We utilise the LiDAR4D scene flow network to enhance temporal consistency in spatio-temporal coordinates via a 4D hybrid multi-resolution hash and plane encoding. This network predicts the prior and post positions of points between adjacent frames and aggregates multi-frame dynamic features for time-consistent reconstruction. During Flow \gls{mlp} training, forward and backward scene flow is predicted using the prior and subsequent frames, capturing changes in point positions across frames. \gls{cd} is used for calculating geometric loss $({\mathcal{L}}_{\mathrm{flow}})$ between the prediction points $\hat{S}$ and ground truth ${S}$ points of the current frame as given by equations \ref{eq: cd scene flow} and \ref{eq: scene flow loss}. It encourages the two adjacent frames of the point cloud transformed by the predicted scene flow to be as consistent as possible.

\begin{equation} \label{eq: cd scene flow}
\mathrm{CD}=\frac{1}{K}\sum_{\hat{p}_{i}\in\hat{S}}\operatorname*{min}_{p_{i}\in S}||\hat{p}_{i}-p_{i}||_{2}^{2}+\frac{1}{K}\sum_{p_{i}\in S}\operatorname*{min}_{\hat{p}_{i}\in\hat{S}}||p_{i}-\hat{p}_{i}||_{2}^{2}
\end{equation}

\begin{equation} \label{eq: scene flow loss}
    {\mathcal{L}}_{\mathrm{flow}}=\sum_{j\in\pm1}{\mathrm{CD}}(S_{i}+{\mathrm{MLP}}_{\mathrm{flow}}(S_{i}),S_{i+j}),i\in(0,n{-}1)
\end{equation}

Total reconstruction of the proposed framework is a combination of weighted six optimization objectives as shown in the Equation \ref{eq: total optimization objectives}. The $\mathcal{L}_{\mathrm{rgb}}$ is derived from Image \gls{nerf} \cite{mildenhall2021nerf} and the rest of the loss function is a combination of LiDAR4D \cite{zheng2024lidar4d} and LiDAR-NeRF \cite{tao2023lidar}.

\begin{equation} \label{eq: total depth loss}
    {\mathcal{L}}_{\mathrm{depth}}=\sum_{\mathbf{r}\in R}\left\|{\hat{D}}(\mathbf{r})-D(\mathbf{r})\right\|_{1}
\end{equation}
\begin{equation} \label{eq: total intensity loss}
    {\mathcal{L}}_{\mathrm{intensity}}=\sum_{\mathbf{r}\in R}\left\|{\hat{I}}(\mathbf{r})-I(\mathbf{r})\right\|_{2}^{2}
\end{equation}
\begin{equation} \label{eq: total raydrop loss}
    {\mathcal{L}}_{\mathrm{raydrop}}=\sum_{\mathbf{r}\in R}\left\|{\hat{P}}(\mathbf{r})-P(\mathbf{r})\right\|_{2}^{2}
\end{equation}
\begin{equation} \label{eq: total rgb loss}
    {\mathcal{L}}_{\mathrm{rgb}}=\sum_{\mathbf{r}\in R}\left\|{\hat{C}}(\mathbf{r})-C(\mathbf{r})\right\|_{2}^{2}
\end{equation}

\begin{equation} \label{eq: total optimization objectives}
\begin{split}
    \mathcal{L}_{total} = & \, \lambda_{1}\mathcal{L}_{\mathrm{flow}}
    +  \lambda_{2}\mathcal{L}_{\mathrm{depth}}(\mathbf{r})
    +  \lambda_{3}\mathcal{L}_{\mathrm{intensity}}(\mathbf{r}) \\
    & + \lambda_{4}\mathcal{L}_{\mathrm{raydrop}}(\mathbf{r})
    +  \lambda_{5}\mathcal{L}_{\mathrm{grad}}    
    + \lambda_{6}\mathcal{L}_{\mathrm{rgb}} 
\end{split}
\end{equation}

where $\mathcal{L}_{\mathrm{depth}}$, $\mathcal{L}_{\mathrm{intensity}}$, $\mathcal{L}_{\mathrm{raydrop}}$ and $\mathcal{L}_{\mathrm{rgb}}$ are losses for Depth, Intensity, Raydrop and RGB of a corresponding pixel. Scene flow loss $\mathcal{L}_{\mathrm{flow}}$ given by the Equation \ref{eq: scene flow loss}. $\lambda_{1}$, $\lambda_{2}$, $\lambda_{3}$, $\lambda_{4}$, $\lambda_{5}$ and $\lambda_{6}$ are  their corresponding weights.  Mean Absolute Error is used for calculating Depth loss while Mean Square Error is used for calculating Intensity loss, Raydrop loss, Gradient loss and RGB loss.

\section{EXPERIMENTS}
\label{sec5}
The proposed \gls{nvsf} is implemented based on LiDAR-NeRF \cite{tao2023lidar} and LiDAR4D \cite{zheng2024lidar4d} codebase. The framework is trained on the KITTI-360 \cite{liao2022kitti} dataset using its Velodyne HDL‐64E \gls{lidar}, two front wide-angle Cameras and ego position data. We train on 4 static and 6 dynamic scenes (same as mentioned in LiDAR4D) each comprising 64 individual frames. Out of this, 60 frames are used for training, and 4 frames are used for validation purposes. The height and width of the \gls{lidar} pano image is set to 66 and 1030. Before training, \gls{lidar} point clouds are centred and scaled such that the region of interest falls within a unit cube. All frames are randomly sampled to prevent over-fitting and promote consistent learning. A ray is sampled uniformly at 768 points in \gls{aabb} frame. The optimization of \gls{nvsf} network is implemented in PyTorch with Adam \cite{kingma2014adam} optimiser. The maximum iterations are set to 30k for each scene, with a batch size of 2048 rays per image. Learning rate is 1e-2 with exponential decay during iterations and a final decay coefficient of 0.1. The values of weights $\lambda_{1}$, $\lambda_{2}$, $\lambda_{3}$, $\lambda_{4}$, $\lambda_{5}$ and $\lambda_{6}$ are 1.0, 1.0, 0.1, 0.01, 0.01 and 1.0 respectively. Remaining parameters such as 4D-Hybrid encoding and U-Net are kept same as the LiDAR4D network. 

\subsection{Baseline}
For a thorough evaluation of the \gls{nvs} framework, a comprehensive comparison is conducted with LiDAR-NeRF\cite{tao2023lidar} and LiDAR4D \cite{zheng2024lidar4d} models on  both static and dynamic scenes reconstruction.  LiDAR4D and LiDAR-NeRF has demonstrated superior performance compared to other LiDAR-based \gls{nvs} methods such as LiDARSim \cite{manivasagam2020lidarsim} and PCGen \cite{allen1980pcgen}. Given their proven efficacy, it is deemed unnecessary to include these other networks in the comparison. The baseline models are trained using their available GitHub code. The primary focus of this work is to synthesise multimodal data though unified framework and introduce novel contributions to enhance their performance.

\begin{table}[!ht]
\setlength{\tabcolsep}{4.5pt}
\centering
    \begin{tabular}{|c|p{1cm}|p{1.1cm}|p{1.1cm}|p{1.1cm}|}
        \hline
        \rule{0pt}{10pt} \multirow{3}{*}{\textbf{Modules}}  & \multicolumn{4}{c|}{\textbf{LiDAR (Foreground)}} \\ \cline{2-5} 
        \rule{0pt}{10pt} & PCD CD $\downarrow$ & Range RMSE $\downarrow$ & Intensity RMSE $\downarrow$ &  Raydrop RMSE $\downarrow$  \\ \hline
        \textbf{BL}& 0.220 & 0.968 & 0.025 & 0.093 \\ 
        \textbf{BL + HS}& 0.173 & 0.846 & 0.023 & 0.080 \\
        \textbf{BL + FA}& 0.158 & \textbf{0.813} & 0.023 & \underline{0.079} \\
        \textbf{BL + ML}& \underline{0.137} & 0.833 & 0.024 & 0.081 \\
        \textbf{NVSF (All)}& \textbf{0.128} & \underline{0.818} & \textbf{0.023} & \textbf{0.078} \\ \hline
        
    \end{tabular}
    \caption{Comparison of LiDAR based metrics for different modules. The ablation is conducted over a subset of scenes. BL, HS, FA and ML is Baseline, Heuristic Sampling, Feature alignment and Multimodal Ray sampling respectively.}
    \label{tab: ablation study}
\end{table}

\begin{table*}[!ht]
	\setlength{\tabcolsep}{3.2pt}
	\begin{tabular}{|*{15}{c|}}
		\hline
		\rule{0pt}{10pt}\multirow{3}{*}{\textbf{Method}} & \multicolumn{13}{c|}{\large LiDAR} \\ \cline{2-14} 
		\rule{0pt}{10pt} & \multicolumn{2}{c|}{\textbf{PCD}} & \multicolumn{4}{c|}{\textbf{Range}} & \multicolumn{4}{c|}{\textbf{Intensity}} & \multicolumn{3}{c|}{\textbf{Raydrop}}  \\ \cline{2-14} 
		\rule{0pt}{10pt} & CD $\downarrow$ & F-score $\uparrow$ & RMSE $\downarrow$ & LPIPS $\downarrow$ & SSIM $\uparrow$ & PSNR $\uparrow$ & RMSE $\downarrow$ & LPIPS $\downarrow$ & SSIM $\uparrow$ & PSNR $\uparrow$ & RMSE $\downarrow$  & Acc.$\uparrow$ & F-score $\uparrow$  \\ \hline
		
		\rule{0pt}{10pt} & \multicolumn{13}{c|}{\textbf{Static + Dynamic Scene}} \\ \hline
		\rule{0pt}{10pt} LiDAR-Nerf \cite{tao2023lidar} & \underline{0.114} & \underline{0.918}  & 3.982   & 0.309 & 0.623 & 26.208 & 0.152  & 0.363   & 0.325  & 16.442 & 0.356 & 0.832 & 0.882 \\
		\rule{0pt}{10pt} LiDAR4D\cite{zheng2024lidar4d} & 0.129 & 0.905  & \underline{3.240} & \underline{0.108} & \underline{0.796} & \underline{28.045} & \underline{0.117} & \underline{0.193} & \underline{0.564} & \underline{18.667} & \underline{0.274} & \underline{0.921} & \underline{0.937}  \\  
		\rule{0pt}{10pt} \gls{nvsf} & \textbf{0.091} & \textbf{0.929} & \textbf{3.150} & \textbf{0.091} & \textbf{0.819} & \textbf{28.322} & \textbf{0.115} & \textbf{0.163} & \textbf{0.579} & \textbf{18.862} & \textbf{0.240} & \textbf{0.925} & \textbf{0.947} \\  \hline
		
		\rule{0pt}{10pt} & \multicolumn{13}{c|}{\textbf{Dynamic Scene (Foreground)}} \\ \hline
		\rule{0pt}{10pt} LiDAR-Nerf \cite{tao2023lidar} & 0.758 & 0.408  & 0.799   & 0.053 & 0.967 & 40.885 & 0.025  & 0.033   & \underline{0.973}  & 32.941 & 0.090 & 0.988 & 0.411 \\ 
		\rule{0pt}{10pt} LiDAR4D \cite{zheng2024lidar4d} & \underline{0.255} & \underline{0.501} & \underline{0.726} & \underline{0.039} & \underline{0.973} & \underline{41.436} & \underline{0.021} & \underline{0.026} & \textbf{0.979} & \underline{34.314} & \underline{0.076} & \underline{0.991} & \underline{0.619}\\ 
		\rule{0pt}{10pt} \gls{nvsf} & \textbf{0.111} & \textbf{0.720}  & \textbf{0.645}   & \textbf{0.030} & \textbf{0.976} & \textbf{42.459} & \textbf{0.020}  & \textbf{0.019}   & \textbf{0.979}  & \textbf{34.870} & \textbf{0.069} & \textbf{0.992} & \textbf{0.733} \\ \hline
	\end{tabular}
	\caption{Quantitative comparison on KITTI 360 Dataset. We report LiDAR based metrics for Static, Dynamic and Foreground (dynamic LiDARPoints) scenes. \textbf{Bold} and \underline{underlined} values are best and second best respectively for the corresponding metrics.}
	\label{tab: results pcd metrics dynamic scenes combined}
\end{table*}

\subsection{LiDAR Evaluation}
We report evaluation metrics for different attributes such as Point Cloud, Range, Intensity, Raydrop and Camera image quality. The metrics are averaged over the 10 scenes, as mentioned above. To analyse dynamic point reconstruction, we report these metrics for the foreground section too. 3D annotation of KITTI360 dataset \cite{liao2022kitti} is used for the foreground separation. The proposed \gls{nvs} framework outperformed both LiDAR-NeRF \cite{tao2023lidar} and LiDAR4D \cite{zheng2024lidar4d} models in almost every PCD synthesis metric. This difference is especially highlighted in the foreground section, as reported in Table \ref{tab: results pcd metrics dynamic scenes combined}. To analyse the effectiveness of individual contributions, we conduct an ablation study as shown in Table \ref{tab: ablation study}. Maximum improvement is shown in the feature alignment, and the efficacy of the multimodal fusion is also evident. Our method has shown that multimodal \gls{nvs} can have a shared \gls{nerf} without impacting individual performance. The framework has shown improvements of 29\% in \gls{cd} and 12\% in Raydrop \gls{rmse}. Higher values of \gls{psnr} suggest that the synthesised point cloud has less noise compared to both baseline models.

\begin{comment}
\begin{table*}[!ht]
\centering
\includegraphics[width=1\textwidth]{Bilder/results_pcd_metrics_dynamic_scenes_combined.png}   
\caption{Table shows the combined (background+foreground) evaluation result for reconstruction of PCD of two Dynamic scenes. \textbf{Bold} and \underline{underlined} values are best and second best respectively for the corresponding metrics.}
\label{tab: results pcd metrics dynamic scenes combined}
\end{table*}
\end{comment}

\begin{comment}
\begin{table*}[!ht]
\centering
\includegraphics[width=1\textwidth]{Bilder/results_pcd_metrics_dynamic_scenes_foreground.png} 
\caption{Table shows the foreground evaluation result for reconstruction of PCD of two Dynamic scenes. \textbf{Bold} and \underline{underlined} values are best and second best respectively for the corresponding metrics.}
\label{tab: results pcd metrics dynamic scenes foreground}
\end{table*}
\end{comment}

\begin{figure}[h]
\centering
\includegraphics[width=0.49\textwidth]{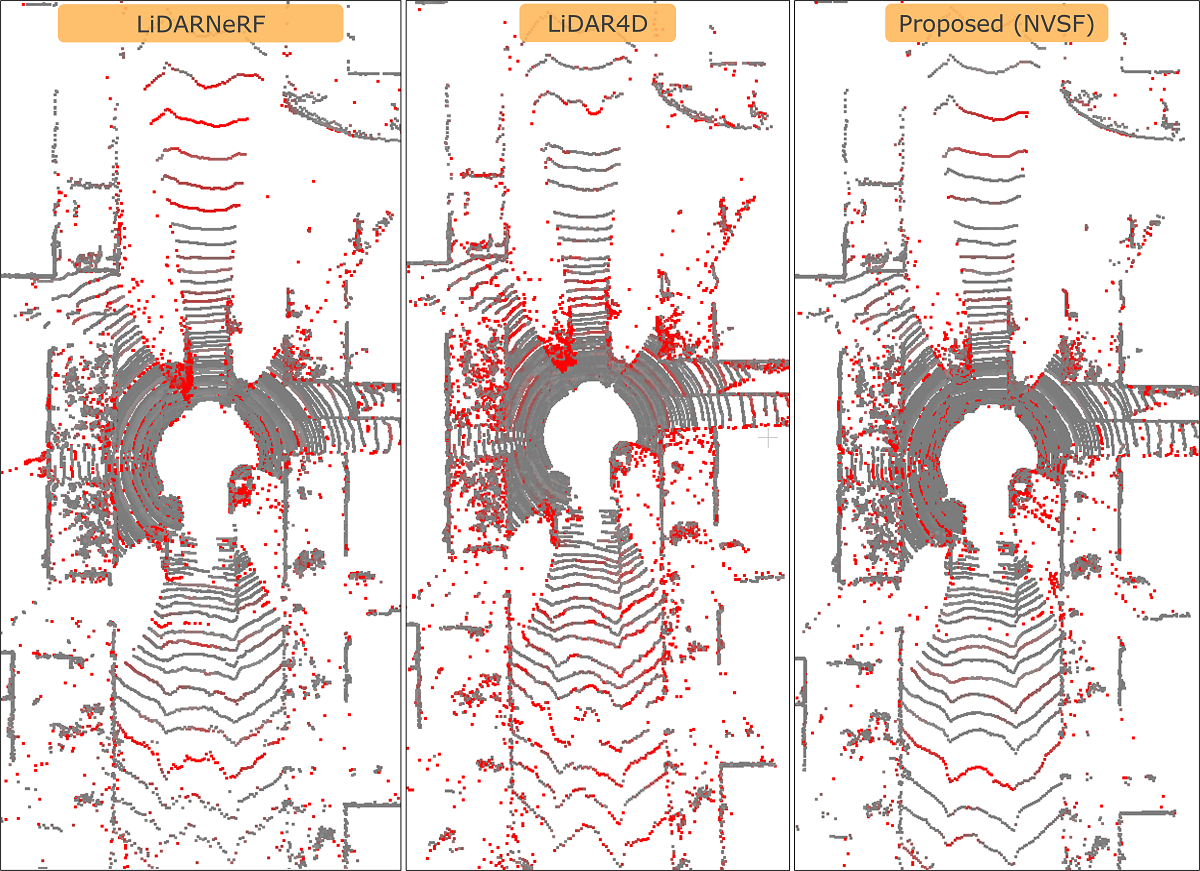}
\caption{Comparison of error map of synthesised PCD by LiDAR-NeRF, LiDAR4D and Proposed NVSF for dynamic sequence 4950. Red colour points are indicating an error of more than 1.0m with respect to ground truth.}
\label{fig: pcd error map comparison for dynamic scene }
\end{figure}

The proposed network has accurately synthesised background and foreground objects with view-dependent attributes compared to baseline and LiDAR4D \cite{zheng2024lidar4d} networks, as shown in Figure \ref{fig: pcd error map comparison for dynamic scene }. Its PCD has better feature representation, consistency and high fidelity.

\subsection{Camera Evaluation}
The proposed \gls{nvs} framework uses multimodal data from \gls{lidar} and Camera for training the network, thereby jointly learning implicit Neural scene representation and view-dependent attributes. The framework can synthesise Camera images along with \gls{lidar} point clouds. Since neither of the Baseline networks support joint synthesis of \gls{lidar} and Camera data, we only report evaluation of our work. Table \ref{tab: results Camera metrics combined and foreground} shows the image quality metrics for \gls{nvsf} model. The combined \gls{psnr} value is low when compared to foreground section, this could be due to the presence of limited dynamic objects. To give a perspective on our quality of image synthesis, NeuRad \cite{tonderski2023neurad} (supervised approach) has reported a \gls{psnr} value of 27 for KITTI MOT dataset \cite{Voigtlaender2019CVPR}. This indicates the effectiveness of self-supervised approach for dynamic object view synthesis.

\begin{comment}
\begin{table}[ht!]
\centering
\includegraphics[width=0.49\textwidth]{Bilder/results_Camera_metrics_combined_and_foreground.png}   
\caption{Table shows the evaluation result for the reconstruction of the Camera image of two Static scenes and two dynamic scenes. Metrics of Dynamic scenes show combined results of static and dynamic objects.}
\label{tab: results Camera metrics combined and foreground}
\end{table}
\end{comment}

\begin{table}[!ht]
\setlength{\tabcolsep}{5pt}
\centering
    \begin{tabular}{|*{5}{c|}}
        \hline
        \rule{0pt}{10pt} \multirow{2}{*}{\textbf{Scenes}}  & \multicolumn{4}{c|}{\textbf{Camera}} \\ \cline{2-5} 
        \rule{0pt}{10pt} & RMSE $\downarrow$ & LPIPS $\downarrow$ & SSIM $\uparrow$ & PSNR $\uparrow$  \\ \hline
        \textbf{Static + Dynamic} & 1.305 & 0.217 & 0.817 & 25.432 \\ 
        \textbf{Dynamic (Foreground)} & 0.344 & 0.017 & 0.736 & 49.057 \\ \hline
        
    \end{tabular}
    \caption{Table shows the evaluation result for the reconstruction of the Camera image for KITTI 360 Dataset.}
    \label{tab: results Camera metrics combined and foreground}
\end{table}

\begin{comment}
\begin{figure}[h!]
\centering
\includegraphics[width=0.49\textwidth]{Bilder/image_comparison_gt_vs_nvsf.png}
\caption{Figure is comparing between Ground Truth and Synthesised image by NVSF. Moving car is shown in red box.}
\label{fig: image comparison gt vs nvsf}
\end{figure}
\end{comment}

% Qualitative analysis
\section{Conclusion}
\label{sec6}
This paper has proposed a \gls{nvs} framework for synthesis of both \gls{lidar} and Camera data by learning implicit Neural representation of a 4D scene and view-dependent attributes in a canonical frame. We report the best performance on quantitative and qualitative metrics when compared to baselines, LiDAR-NeRF and LiDAR4D. We revisit the fundamentals of ray sampling and improve upon this by introducing Heuristic sampling and Double gradient mask. The proposed network is able to synthesise both static and dynamic scenes in a self-supervised manner. Despite the performance, the current self-supervision approach for dynamic scene reconstruction binds foreground and background scenes, which restrict \gls{nvs} within the ego-vehicle trajectory in the spatio-temporal coordinate space. Consequently, scene editing or scene manipulation is not possible outside ego-vehicle trajectory. We foresee the future multimodal work will focus on improving the reconstruction quality over larger scenes without 3D annotation supervision. 
% The proposed framework synthesizes both LiDAR point clouds and Camera images, simulated data can narrow the domain gaps as demonstrated in the the downstream task.  However, there is a noticeable gap in AP values of original \gls{pcd} and synthesized \gls{pcd}, attributed to the sim2real and inter dataset domain gaps.

\section{Acknowledgement}
\label{sec7}
This work is a part of a master thesis supported by AVL Software and Functions GmbH, Germany and Technische Hochschule Deggendorf, Germany. We are thankful to publishers of LiDAR-NeRF and LiDAR4D papers for their great work and making the codebase publicly available for further research. 

\newpage

\bibliographystyle{IEEETran.bst}
\bibliography{IEEEabrv}

\end{document}

%% file: glossaries.tex
%%%%%%%%%%%%%%%%%%%%%%%%%%%%%%%%%%%%%%%%%%%%%%%%%%%%%%%%%%
%
% Nomenklatur, Symbolverzeichnis
%
%%%%%%%%%%%%%%%%%%%%%%%%%%%%%%%%%%%%%%%%%%%%%%%%%%%%%%%%%%

% General
% \newacronym{ad}{AD}{Autonomous Driving}
\newacronym{nvsf}{NVSF}{Novel View Synthesis Framework}
\newacronym{av}{AV}{Autonomous Vehicle}
\newacronym{ai}{AI}{Artificial Intelligence}
\newacronym{ml}{ML}{Machine Learning}
\newacronym{sota}{SOTA}{State of the Art}
\newacronym{avl}{AVL}{AVL Software and Functions GmbH}
\newacronym{adas}{ADAS}{Advanced Driver Assistance Systems}
\newacronym{aiv}{AIV}{Autonomous Intelligent Vehicles}
\newacronym{dlt}{DLT}{Direct Linear Transform}
\newacronym{ec}{EC}{Extrinsic Calibration}
% Sensors and data
\newacronym{gps}{GPS}{Global Positioning System}
\newacronym{imu}{IMU}{Inertia Measurement Unit}
\newacronym{gnss}{GNSS}{Global Navigation Satellite System}
\newacronym{lidar}{LiDAR}{Light Detection and Ranging}
\newacronym{pcd}{PCD}{Point cloud data}
\newacronym{slam}{SLAM}{Simultaneous Localization and Mapping}
\newacronym{fov}{FoV}{Field of View}
\newacronym{vfov}{vFoV}{Vertical Field of View}
\newacronym{hfov}{hFoV}{Horizontal Field of View}

% Neural Network
\newacronym{ann}{ANN}{Artificial Neural Network}
\newacronym{mlp}{MLP}{Multi Layer Perceptron}
\newacronym{bbox}{BBox}{Bounding Box}
\newacronym{nerf}{NeRF}{Neural Radiance Fields}
\newacronym{nvs}{NVS}{Novel View Synthesis}
\newacronym{nsg}{NSG}{Neural Scene Graph}
\newacronym{cnn}{CNN}{Convolutional Neural Network} 
\newacronym{fcn}{FCN}{Fully Connected Network} 
\newacronym{pdf}{PDF}{Probability Density Function} 
\newacronym{cdf}{CDF}{Cumulative Density Function} 
\newacronym{hd}{HD}{High Definition}
\newacronym{rgb}{RGB}{Red-Green-Blue color code}
\newacronym{iou}{IoU}{Intersection over Union}  
\newacronym{nms}{NMS}{Non-Max Supression}  
\newacronym{ssd}{SSD}{Single Shot Detector}  
\newacronym{aabb}{AABB}{Axis Aligned Bounding Box}  
%matrices and loss functions
\newacronym{mae}{MAE}{Mean Absolute Error}  
\newacronym{medae}{MedAE}{Median Absolute Error}  
\newacronym{mse}{MSE}{Mean Square Error}  
\newacronym{rmse}{RMSE}{Root Mean Square Error}  
\newacronym{psnr}{PSNR}{Peak Signal to noise ratio}  
\newacronym{ssim}{SSIM}{Structural Similarity Index}  
\newacronym{lpips}{LPIPS}{Learned Perceptual Image Patch Similarity}  
\newacronym{tp}{TP}{True Positive}  
\newacronym{fp}{FP}{Flase Positive}  
\newacronym{fn}{FN}{Flase Negative}  
\newacronym{ap}{AP}{Average Precision}  
\newacronym{map}{mAP}{Mean Average Precision}
\newacronym{cd}{CD}{Chamfer distance}
\newacronym{emd}{EMD}{Earth Mover's Distance}
\newacronym{bce}{BCE}{Binary Cross Entropy}
\newacronym{mrhe}{MRHE}{Multi-resolution hash encoding}

%misc
\newacronym{dgt}{DGT}{Dynamic Ground Truth}
%\setglossarystyle{long2col}

\newacronym{fmcw}{FMCW}{Frequency-Modulated Continuous-Wave}
\newacronym{cfar}{CFAR}{Constant False Alarm Rate}
\newacronym{rcs}{RCS}{Radar Cross Section}
\newacronym{dnn}{DNN}{Deep Neural Network}
\newacronym{stft}{STFT}{Short Time Fourier Transformation}
\newacronym{rd}{RD}{Range-Doppler}
\newacronym{roi}{ROI}{Region-of-Interest}
\newacronym{yolo}{YOLO}{You Only Look Once}
\newacronym{ra}{RA}{Range-Angle}
\newacronym{ad}{AD}{Angle-Doppler}
\newacronym{rad}{RAD}{Range-Angle-Doppler}
\newacronym{nn}{NN}{Neural Network}
\newacronym{bb}{BB}{Bounding Box}
\newacronym{dnms}{DNMS}{Distance-based non-max Suppression}
\newacronym{vru}{VRU}{Vulnerable Road User}
\newacronym{bev}{BEV}{Bird's-eye view}
\newacronym{cdc}{CDC}{Convolution-deconvolution}
\newacronym{fft}{FFT}{Fast Fourier Transform}
\newacronym{soc}{SOC}{System on Chip}

%% file: conference_101719.bbl
\begin{thebibliography}{10}
\providecommand{\url}[1]{#1}
\csname url@rmstyle\endcsname
\providecommand{\newblock}{\relax}
\providecommand{\bibinfo}[2]{#2}
\providecommand\BIBentrySTDinterwordspacing{\spaceskip=0pt\relax}
\providecommand\BIBentryALTinterwordstretchfactor{4}
\providecommand\BIBentryALTinterwordspacing{\spaceskip=\fontdimen2\font plus
\BIBentryALTinterwordstretchfactor\fontdimen3\font minus \fontdimen4\font\relax}
\providecommand\BIBforeignlanguage[2]{{%
\expandafter\ifx\csname l@#1\endcsname\relax
\typeout{** WARNING: IEEEtran.bst: No hyphenation pattern has been}%
\typeout{** loaded for the language `#1'. Using the pattern for}%
\typeout{** the default language instead.}%
\else
\language=\csname l@#1\endcsname
\fi
#2}}

\bibitem{dosovitskiy2017carla}
A.~Dosovitskiy, G.~Ros, F.~Codevilla, A.~Lopez, and V.~Koltun, ``{CARLA}: An open urban driving simulator,'' in \emph{Conference on robot learning}.\hskip 1em plus 0.5em minus 0.4em\relax PMLR, 2017, pp. 1--16.

\bibitem{manivasagam2020lidarsim}
S.~Manivasagam, S.~Wang, K.~Wong, W.~Zeng, M.~Sazanovich, S.~Tan, B.~Yang, W.-C. Ma, and R.~Urtasun, ``{LiDARsim: Realistic LiDAR} simulation by leveraging the real world,'' in \emph{Proceedings of the IEEE/CVF Conference on Computer Vision and Pattern Recognition}, 2020, pp. 11\,167--11\,176.

\bibitem{mildenhall2021nerf}
B.~Mildenhall, P.~P. Srinivasan, M.~Tancik, J.~T. Barron, R.~Ramamoorthi, and R.~Ng, ``{NeRF}: Representing scenes as neural radiance fields for view synthesis,'' \emph{Communications of the ACM}, vol.~65, no.~1, pp. 99--106, 2021.

\bibitem{zheng2024lidar4d}
Z.~Zheng, F.~Lu, W.~Xue, G.~Chen, and C.~Jiang, ``{LiDAR4D: Dynamic Neural Fields for Novel Space-time View LiDAR Synthesis},'' \emph{arXiv preprint arXiv:2404.02742}, 2024.

\bibitem{fridovich2023k}
S.~Fridovich-Keil, G.~Meanti, F.~R. Warburg, B.~Recht, and A.~Kanazawa, ``K-planes: {Explicit radiance fields} in space, time, and appearance,'' in \emph{Proceedings of the IEEE/CVF Conference on Computer Vision and Pattern Recognition}, 2023, pp. 12\,479--12\,488.

\bibitem{ost2021neural}
J.~Ost, F.~Mannan, N.~Thuerey, J.~Knodt, and F.~Heide, ``{Neural Scene Graphs for Dynamic Scenes},'' in \emph{Proceedings of the IEEE/CVF Conference on Computer Vision and Pattern Recognition}, 2021, pp. 2856--2865.

\bibitem{barron2022mip}
J.~T. Barron, B.~Mildenhall, D.~Verbin, P.~P. Srinivasan, and P.~Hedman, ``{Mip-NeRF} 360: Unbounded anti-aliased neural radiance fields,'' in \emph{Proceedings of the IEEE/CVF Conference on Computer Vision and Pattern Recognition}, 2022, pp. 5470--5479.

\bibitem{barron2023zip}
------, ``{Zip-NeRF}: Anti-aliased grid-based neural radiance fields,'' in \emph{Proceedings of the IEEE/CVF International Conference on Computer Vision}, 2023, pp. 19\,697--19\,705.

\bibitem{muller2022instant}
T.~M{\"u}ller, A.~Evans, C.~Schied, and A.~Keller, ``Instant neural graphics primitives with a multiresolution hash encoding,'' \emph{ACM transactions on graphics (TOG)}, vol.~41, no.~4, pp. 1--15, 2022.

\bibitem{chen2022tensorf}
A.~Chen, Z.~Xu, A.~Geiger, J.~Yu, and H.~Su, ``Tensorf: Tensorial radiance fields,'' in \emph{European conference on computer vision}.\hskip 1em plus 0.5em minus 0.4em\relax Springer, 2022, pp. 333--350.

\bibitem{fridovich2022plenoxels}
S.~Fridovich-Keil, A.~Yu, M.~Tancik, Q.~Chen, B.~Recht, and A.~Kanazawa, ``Plenoxels: Radiance fields without neural networks,'' in \emph{Proceedings of the IEEE/CVF conference on computer vision and pattern recognition}, 2022, pp. 5501--5510.

\bibitem{sun2022direct}
C.~Sun, M.~Sun, and H.-T. Chen, ``Direct voxel grid optimization: Super-fast convergence for radiance fields reconstruction,'' in \emph{Proceedings of the IEEE/CVF conference on computer vision and pattern recognition}, 2022, pp. 5459--5469.

\bibitem{chan2022efficient}
E.~R. Chan, C.~Z. Lin, M.~A. Chan, K.~Nagano, B.~Pan, S.~De~Mello, O.~Gallo, L.~J. Guibas, J.~Tremblay, S.~Khamis, \emph{et~al.}, ``Efficient geometry-aware 3d generative adversarial networks,'' in \emph{Proceedings of the IEEE/CVF conference on computer vision and pattern recognition}, 2022, pp. 16\,123--16\,133.

\bibitem{hu2023tri}
W.~Hu, Y.~Wang, L.~Ma, B.~Yang, L.~Gao, X.~Liu, and Y.~Ma, ``{Tri-MipRF}: {Tri-Mip} representation for efficient anti-aliasing neural radiance fields,'' in \emph{Proceedings of the IEEE/CVF International Conference on Computer Vision}, 2023, pp. 19\,774--19\,783.

\bibitem{sun2024dil}
S.~Sun, B.~Zhuang, Z.~Jiang, B.~Liu, X.~Xie, and M.~Chandraker, ``{LidaRF}: Delving into {LiDAR} for neural radiance field on street scenes,'' \emph{arXiv preprint arXiv:2405.00900}, 2024.

\bibitem{tancik2022block}
M.~Tancik, V.~Casser, X.~Yan, S.~Pradhan, B.~Mildenhall, P.~P. Srinivasan, J.~T. Barron, and H.~Kretzschmar, ``{Block-NeRF}: Scalable large scene neural view synthesis,'' in \emph{Proceedings of the IEEE/CVF Conference on Computer Vision and Pattern Recognition}, 2022, pp. 8248--8258.

\bibitem{tao2023lidar}
T.~Tao, L.~Gao, G.~Wang, Y.~Lao, P.~Chen, H.~Zhao, D.~Hao, X.~Liang, M.~Salzmann, and K.~Yu, ``{LiDAR-NeRF}: {Novel LiDAR} view synthesis via neural radiance fields,'' \emph{arXiv preprint arXiv:2304.10406}, 2023.

\bibitem{zhang2023nerf}
J.~Zhang, F.~Zhang, S.~Kuang, and L.~Zhang, ``{NeRF-LiDAR}: {Generating Realistic LiDAR Point Clouds with Neural Radiance Fields},'' in \emph{AAAI Conference on Artificial Intelligence (AAAI)}, 2024.

\bibitem{rematas2022urban}
K.~Rematas, A.~Liu, P.~P. Srinivasan, J.~T. Barron, A.~Tagliasacchi, T.~Funkhouser, and V.~Ferrari, ``Urban radiance fields,'' in \emph{Proceedings of the IEEE/CVF Conference on Computer Vision and Pattern Recognition}, 2022, pp. 12\,932--12\,942.

\bibitem{wang2023neural}
Z.~Wang, T.~Shen, J.~Gao, S.~Huang, J.~Munkberg, J.~Hasselgren, Z.~Gojcic, W.~Chen, and S.~Fidler, ``Neural fields meet explicit geometric representations for inverse rendering of urban scenes,'' in \emph{Proceedings of the IEEE/CVF Conference on Computer Vision and Pattern Recognition}, 2023, pp. 8370--8380.

\bibitem{tonderski2023neurad}
A.~Tonderski, C.~Lindstr{\"o}m, G.~Hess, W.~Ljungbergh, L.~Svensson, and C.~Petersson, ``{NeuRAD: Neural} rendering for autonomous driving,'' \emph{arXiv preprint arXiv:2311.15260}, 2023.

\bibitem{yang2023unisim}
Z.~Yang, Y.~Chen, J.~Wang, S.~Manivasagam, W.-C. Ma, A.~J. Yang, and R.~Urtasun, ``{UniSim: A Neural Closed-Loop Sensor Simulator},'' in \emph{CVPR}, 2023.

\bibitem{huang2023neural}
S.~Huang, Z.~Gojcic, Z.~Wang, F.~Williams, Y.~Kasten, S.~Fidler, K.~Schindler, and O.~Litany, ``{Neural LiDAR} fields for novel view synthesis,'' in \emph{Proceedings of the IEEE/CVF International Conference on Computer Vision}, 2023, pp. 18\,236--18\,246.

\bibitem{zhang2024nerf}
J.~Zhang, F.~Zhang, S.~Kuang, and L.~Zhang, ``{NeRF-LiDAR}: Generating realistic {LiDAR} point clouds with neural radiance fields,'' in \emph{Proceedings of the AAAI Conference on Artificial Intelligence}, vol.~38, no.~7, 2024, pp. 7178--7186.

\bibitem{NEURIPS2021_41263b9a}
\BIBentryALTinterwordspacing
X.~Li, J.~Kaesemodel~Pontes, and S.~Lucey, ``{Neural Scene Flow Prior},'' in \emph{Advances in Neural Information Processing Systems}, M.~Ranzato, A.~Beygelzimer, Y.~Dauphin, P.~Liang, and J.~W. Vaughan, Eds., vol.~34.\hskip 1em plus 0.5em minus 0.4em\relax Curran Associates, Inc., 2021, pp. 7838--7851. [Online]. Available: \url{https://proceedings.neurips.cc/paper_files/paper/2021/file/41263b9a46f6f8f22668476661614478-Paper.pdf}
\BIBentrySTDinterwordspacing

\bibitem{wu2023mars}
Z.~Wu, T.~Liu, L.~Luo, Z.~Zhong, J.~Chen, H.~Xiao, C.~Hou, H.~Lou, Y.~Chen, R.~Yang, \emph{et~al.}, ``{MARS}: An instance-aware, modular and realistic simulator for autonomous driving,'' in \emph{CAAI International Conference on Artificial Intelligence}.\hskip 1em plus 0.5em minus 0.4em\relax Springer, 2023, pp. 3--15.

\bibitem{otonari2022non}
T.~Otonari, S.~Ikehata, and K.~Aizawa, ``Non-uniform sampling strategies for {NeRF} on 360° images.'' in \emph{BMVC}, 2022, p. 344.

\bibitem{depthnerf}
K.~Deng, A.~Liu, J.-Y. Zhu, and D.~Ramanan, ``Depth-supervised {NeRF}: {Fewer Views and Faster Training for Free},'' in \emph{2022 IEEE/CVF Conference on Computer Vision and Pattern Recognition (CVPR)}, 2022, pp. 12\,872--12\,881.

\bibitem{tang2024alignmif}
T.~Tang, G.~Wang, Y.~Lao, P.~Chen, J.~Liu, L.~Lin, K.~Yu, and X.~Liang, ``{AlignMiF: Geometry-Aligned Multimodal Implicit Field for LiDAR-Camera Joint Synthesis},'' \emph{arXiv preprint arXiv:2402.17483}, 2024.

\bibitem{peng2022balanced}
X.~Peng, Y.~Wei, A.~Deng, D.~Wang, and D.~Hu, ``Balanced multimodal learning via on-the-fly gradient modulation,'' in \emph{Proceedings of the IEEE/CVF conference on computer vision and pattern recognition}, 2022, pp. 8238--8247.

\bibitem{sun2021learning}
Y.~Sun, S.~Mai, and H.~Hu, ``Learning to balance the learning rates between various modalities via adaptive tracking factor,'' \emph{IEEE Signal Processing Letters}, vol.~28, pp. 1650--1654, 2021.

\bibitem{wang2020makes}
W.~Wang, D.~Tran, and M.~Feiszli, ``What makes training multi-modal classification networks hard?'' in \emph{Proceedings of the IEEE/CVF conference on computer vision and pattern recognition}, 2020, pp. 12\,695--12\,705.

\bibitem{li2021neural}
X.~Li, J.~Kaesemodel~Pontes, and S.~Lucey, ``Neural scene flow prior,'' \emph{Advances in Neural Information Processing Systems}, vol.~34, pp. 7838--7851, 2021.

\bibitem{Li_2023_ICCV}
X.~Li, J.~Zheng, F.~Ferroni, J.~K. Pontes, and S.~Lucey, ``{Fast Neural Scene Flow},'' in \emph{Proceedings of the IEEE/CVF International Conference on Computer Vision (ICCV)}, October 2023, pp. 9878--9890.

\bibitem{zheng2023neuralpci}
Z.~Zheng, D.~Wu, R.~Lu, F.~Lu, G.~Chen, and C.~Jiang, ``{NeuralPCI: Spatio-temporal Neural Field for 3D Point Cloud Multi-frame Non-linear Interpolation},'' in \emph{Proceedings of the IEEE/CVF Conference on Computer Vision and Pattern Recognition}, 2023, pp. 909--918.

\bibitem{li2023dynibar}
Z.~Li, Q.~Wang, F.~Cole, R.~Tucker, and N.~Snavely, ``{DynIBaR: Neural} dynamic image-based rendering,'' in \emph{Proceedings of the IEEE/CVF Conference on Computer Vision and Pattern Recognition}, 2023, pp. 4273--4284.

\bibitem{liao2022kitti}
Y.~Liao, J.~Xie, and A.~Geiger, ``{KITTI-360}: A novel dataset and benchmarks for urban scene understanding in 2d and 3d,'' \emph{IEEE Transactions on Pattern Analysis and Machine Intelligence}, vol.~45, no.~3, pp. 3292--3310, 2022.

\bibitem{kingma2014adam}
D.~P. Kingma and J.~Ba, ``Adam: A method for stochastic optimization,'' \emph{arXiv preprint arXiv:1412.6980}, 2014.

\bibitem{allen1980pcgen}
G.~A. Allen and L.~M. Baldwin, ``{PCGen}: A {FORTRAN IV} program to generate paired-comparison stimuli,'' \emph{Behavior Research Methods \& Instrumentation}, vol.~12, no.~3, pp. 383--384, 1980.

\bibitem{Voigtlaender2019CVPR}
P.~Voigtlaender, M.~Krause, A.~Osep, J.~Luiten, B.~B.~G. Sekar, A.~Geiger, and B.~Leibe, ``{MOTS: Multi-Object Tracking and Segmentation},'' in \emph{Conference on Computer Vision and Pattern Recognition (CVPR)}, 2019.

\end{thebibliography}
